\documentclass[runningheads]{llncs}
\usepackage[T1]{fontenc}

\usepackage{graphicx,verbatim}
\usepackage{amsmath}
\usepackage{amssymb}
\usepackage{booktabs}
\usepackage{multirow} 
\usepackage{array}

\newcommand{\figref}[1]{Fig. \ref{#1}}

\newcommand{\mymodel}{SurgLQA}
\usepackage[colorlinks=true, citecolor=blue, linkcolor=blue]{hyperref}
\usepackage{marvosym}

\def\eg{\emph{e.g.}}
\def\ie{\emph{i.e.}}

\begin{document}
\title{SurgLQA: Scalable Long-Horizon Surgical Video Question Answering}
%
\author{Diandian Guo, Xikai Yang, Ruiyang Li, Jialun Pei\,$^{\textrm{\Letter}}$, Pheng-Ann Heng}
\institute{The Chinese University of Hong Kong \\
\email{jialunpei@cuhk.edu.hk}}
\authorrunning{Diandian Guo et al.}

  
\maketitle           
\begin{abstract}

Surgical Video Question Answering (VideoQA) provides a promising paradigm for dynamic intraoperative interpretation, enabling real-time decision support and context-aware retrieval in clinical environments. Nevertheless, existing approaches are predominantly restricted to images or short clips, limiting their ability to model long-range procedural dynamics and causal dependencies across extended surgical workflows. To address this challenge, we propose SurgLQA, a unified long-horizon VideoQA framework for scalable surgical reasoning. This framework incorporates Faithful Temporal Consolidation (FTC), which leverages intrinsic temporal cues to construct compact long-range representations while preserving fine-grained temporal fidelity. Further, we develop Temporally-Grounded Multi-Policy Scaling (TMS), an adaptive test-time inference paradigm that strategically adjusts policy-level reasoning capacity within temporally grounded contexts. To facilitate systematic evaluation, we restructured a long-duration colonoscopy VideoQA benchmark, Colon-LQA, and conducted extensive experiments on Colon-LQA and REAL-Colon-VQA. Experimental results demonstrate that our approach achieves consistent performance gains in long-range reasoning with temporally grounded inference. Code link: \href{https://github.com/RascalGdd/SurgLQA}{SurgLQA}.


\end{abstract}
\section{Introduction}
With modern surgical procedures becoming increasingly complex and prolonged, clinical environments demand a new paradigm capable of dynamically understanding and reasoning across the entire surgical process~\cite{varghese2024artificial}, rather than relying solely on stage-specific analysis of predefined objectives~\cite{Cao2023IntelligentWorkflow,skit}.
Cognitive decision-making in clinical settings frequently relies on cross-stage contextual information and a holistic understanding of the evolution of surgical procedures~\cite{pei2025instrument,s2former}. In this context, surgical Video Question Answering (VideoQA) has emerged as a promising paradigm by integrating natural language with surgical video interactions, thereby facilitating flexible workflow reasoning within a unified inference framework~\cite{Seenivasan2023SurgicalGPT}.
This technology not only provides real-time intraoperative decision assistance~\cite{lovit,pmlr50} but also supports postoperative review, surgical quality assessment, and data-driven clinical research~\cite{Qiu2025LLMClinicalReasoning,unisurg,bleeding,surgvista}.

However, scalable surgical VideoQA remains fundamentally constrained by long-horizon temporal reasoning. Unlike general video understanding tasks, surgical procedures often span tens of minutes to hours, where clinically critical events occur sparsely and depend on long-range causal relationships across different operation stages. Consequently, effective surgical VideoQA needs to meet the following requirements: 1) precise temporal modeling across extended durations, 2) accurate localization of sparse yet critical surgical events, and 3) scalable collaborative reasoning under limited computational resources. 

\begin{figure}[t]
    \centering
    \begin{tabular}{c}
        \includegraphics[width=1.00\textwidth]{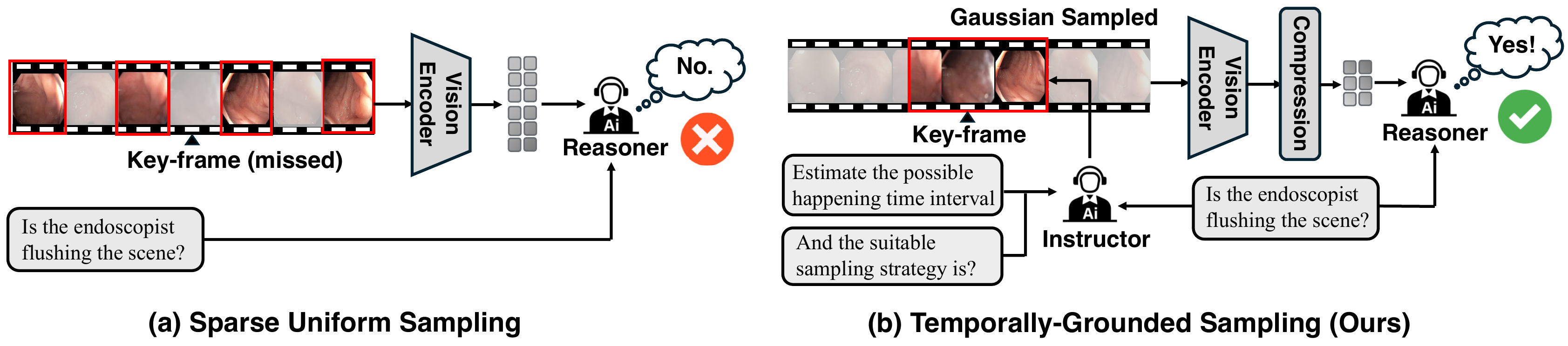} 
    \end{tabular}%
    \caption{(a) Existing VideoQA methods directly encode uniformly sampled frames, which may overlook subtle temporal evidence in keyframes and incur increased computational overhead; (b) \mymodel~adopt a temporally grounded sampling policy and compression mechanism to construct focused video representations.}
\label{fig:first} 
\end{figure}

Existing surgical VQA research faces two intertwined limitations in both benchmarks and modeling paradigms~\cite{pitvqa,surgvivqa,kvasir-vqa}.
On the data side, most existing benchmarks emphasize image-level or short clip–level settings~\cite{ssgvqa,surgvivqa}, prioritizing static visual cues such as instrument type or spatial layout. Corresponding question–answer pairs are typically anchored to narrowly defined frames~\cite{pitvqa,surgicalvqa} or brief video segments~\cite{surgvivqa}, diverging from real clinical workflows where reasoning involves cross-episode evidence aggregation over extended temporal horizons. 
On the modeling side, most surgical vision–language models rely on fixed uniform or sparse sampling strategies~\cite{qwen,surgvivqa} despite the inherently heterogeneous dynamics of surgical workflows. Consequently, such strategies may overlook sparsely distributed yet clinically critical events (see Fig.~\ref{fig:first}), undermine question-conditioned temporal reasoning, and compromise scalability for long videos.
Collectively, these limitations leave long surgical VideoQA largely unexplored.

In light of these challenges, we propose \mymodel, a unified surgical VideoQA framework for scalable and temporally grounded procedural reasoning. \mymodel~models long-duration surgical workflows through \textit{Faithful Temporal Consolidation} (FTC), which leverages intrinsic temporal cues to compress frame-level representations into compact yet temporally faithful embeddings.
Building upon this representation, we introduce \textit{Temporally-Grounded Multi-Policy Scaling} (TMS), an adaptive inference mechanism that dynamically modulates reasoning granularity across temporally localized contexts.
As illustrated in Fig.~\ref{fig:first}, TMS embraces temporally grounded adaptive reasoning instead of sparse uniform sampling, enabling dynamic concentrate on clinically relevant segments to achieve scalable and reliable long-horizon surgical VideoQA.
Additionally, we restructure a long-duration colonoscopy VideoQA benchmark, Colon-LQA, by concatenating multiple temporally ordered real video segments with corresponding question–answer pairs to construct extended sequences~\cite{videoweave}. 
Extensive experiments on Colon-LQA and REAL-Colon-VQA~\cite{surgvivqa} demonstrate that~\mymodel~improves long-range reasoning through event-level temporal grounding.

\section{Methodology}

\subsection{Framework Overview}
\begin{figure}[t]
    \centering
    \begin{tabular}{c}
        \includegraphics[width=0.99\textwidth]{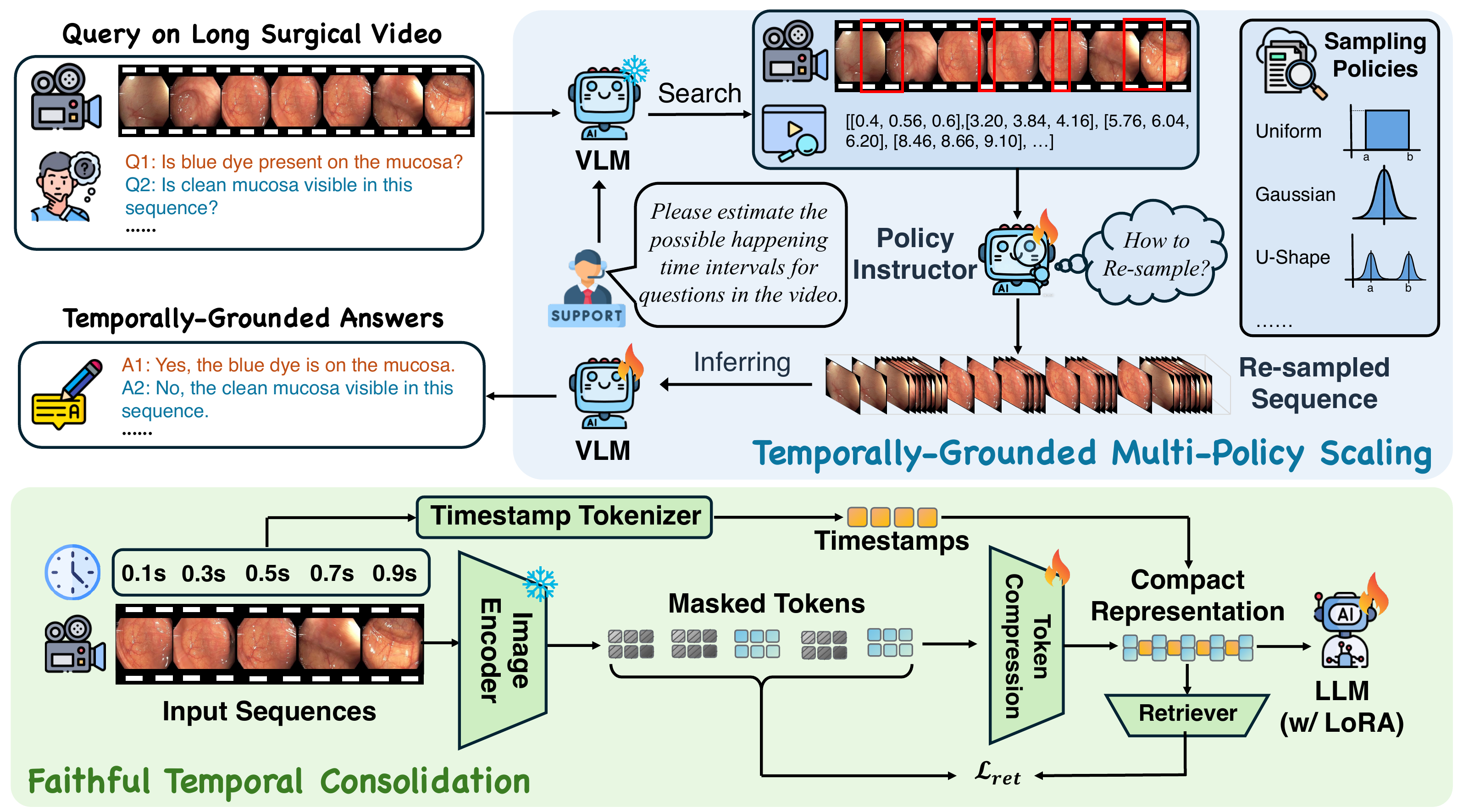} 
    \end{tabular}
    \caption{Overview of the proposed framework for long surgical VideoQA.}

\label{fig:overview} 
\end{figure}

Fig.~\ref{fig:overview} illustrates the proposed framework for long surgical VideoQA.
We first employ \emph{Faithful Temporal Consolidation} (FTC) to compress long video streams into compact, temporally structured representations for efficient long-horizon modeling.
Next, we apply \emph{Temporally-Grounded Multi-Policy Scaling} (TMS) to perform query-conditioned temporal grounding. Specifically, TMS identifies candidate intervals relevant to the query and selects an appropriate sampling policy for each grounded window, generating a query-adaptive re-sampled sequence with refined temporal resolution.
The re-sampled sequence is then directly fed into the vision–language model for the final answer generation.

\subsection{Faithful Temporal Consolidation}

Typically long surgical videos generate massive visual tokens, rendering exhaustive encoding computationally prohibitive.
To this end, we propose \textit{Faithful Temporal Consolidation} (FTC), a learnable spatiotemporal token compression module that reduces redundancy while preserving temporal fidelity.
Let an input video be denoted as 
$\mathbf{V} = \{\mathbf{I}_t\}_{t=1}^{T}$, 
$\mathbf{I}_t \in \mathbb{R}^{C \times H \times W}$. 
The video is partitioned into non-overlapping 3D blocks of size 
$p_t \times p_s \times p_s$, yielding

\begin{equation}
{\small
\mathbf{V}' \in \mathbb{R}^{N \times C \times p_t \times p_s \times p_s}, 
\quad 
N = \frac{T}{p_t}\frac{H}{p_s}\frac{W}{p_s}.
}
\end{equation}
Then, we apply a stack of temporal-aware 3D convolution layers with SiLU activation to the sequence $\mathbf{V}'$, encouraging the model to learn compact and informative spatiotemporal representations $\{\mathbf{Z}_l\}$:
\begin{equation}
{\small
\mathbf{Z}_l = \sigma\!\left(\mathrm{Conv3D}_l(\mathbf{Z}_{l-1})\right), 
\quad l=1,\dots,L.
}
\end{equation}
where $\sigma(\cdot)$ denotes the SiLU activation function.
To preserve explicit temporal cues for long-horizon reasoning, we adopt an interleaved timestamp-token design, constructing a text prompt
$\tau_i=\texttt{``timestamp: }t_i\texttt{ seconds''}$ and tokenize it as $T_i$. 
The sequence is formed by interleaving timestamp and visual tokens:
\begin{equation}
{\small
S = [T_1; V_1; T_2; V_2; \cdots; T_{N'}; V_{N'}],\quad T_i = \phi_{\text{tokenizer}}(\tau_i),
}
\end{equation}
where $V_i$ denotes the consolidated tokens of the $i$-th temporal unit.
This explicit temporal marking can retrieve temporal boundaries in long videos by reading inserted timestamp tokens.
Besides, we bring an auxiliary retrieving loss
$\mathcal{L}_{\mathrm{ret}}$ to preserve pre-compression fidelity.
Giving pre-compression tokens $\mathbf{Z}_1$ and consolidated representations $\mathbf{Z}_{\mathrm{FTC}}$, we randomly sample an index set $\mathcal{M}$$\subset$$\{1,\dots,N\}$.
To prevent trivial copying from temporally adjacent tokens, random masking is applied to the compressed representation,
followed by a lightweight retriever $g(\cdot)$ to predict masked tokens. The retrieving objective is formulated as

\begin{equation}
{\small
\mathcal{L}_{\mathrm{ret}}
=
\sum_{i \in \mathcal{M}}
\left\|
\big[
g(\mathrm{Mask}(\mathbf{Z}_{\mathrm{FTC}}))
\big]_i
-
\mathbf{z}_i^{(1)}
\right\|_2^2,
}
\end{equation}
where $\mathbf{z}_i^{(1)}$ denotes the $i$-th pre-compression token for retrieving.
By reconstructing masked intermediate features, FTC encourages the compressed representation to preserve temporal fidelity under token compression.

\begin{figure}[t]
    \centering
    \begin{tabular}{c}
        \includegraphics[width=0.98\textwidth]{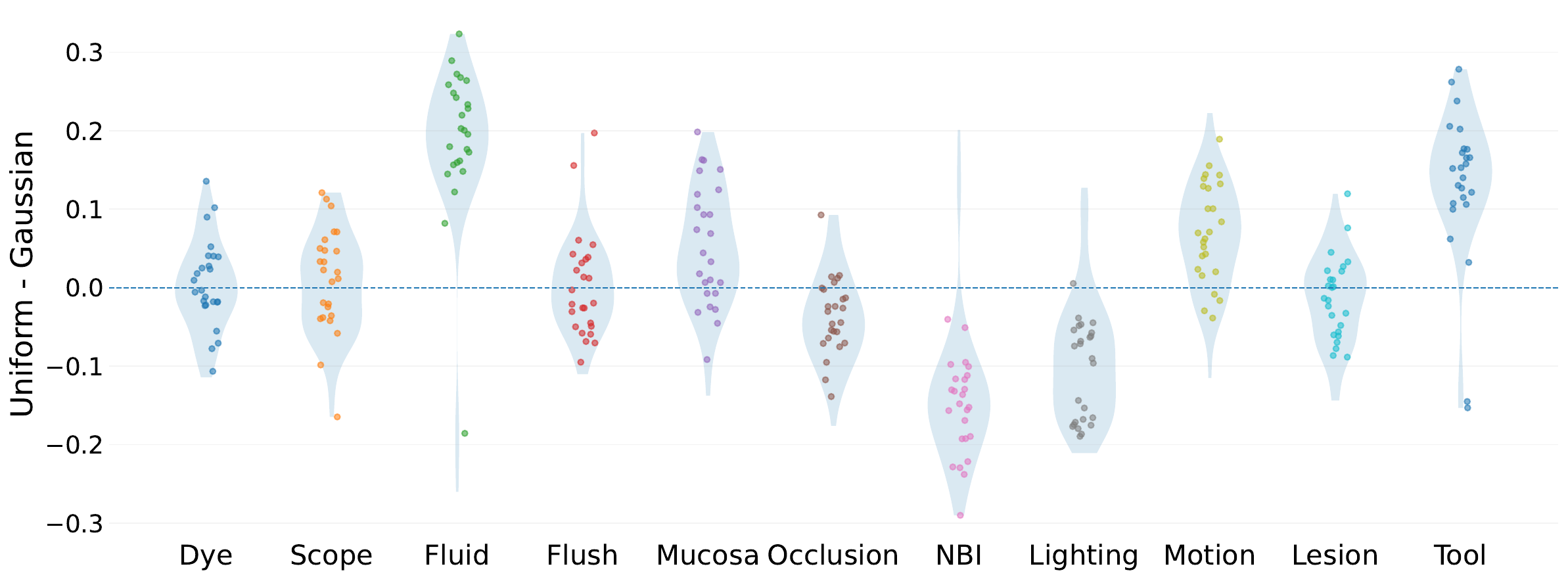} 
    \end{tabular}
    \caption{Instructor reveals distinct sampling preferences across question types. Temporally fine-grained events (\eg, fluid, motion) favor Gaussian sampling, while global static questions (\eg, lighting) benefit more from uniform sampling.}

\label{fig:vis_policy} 
\end{figure}

\subsection{Temporally-Grounded Multi-Policy Scaling}

Current video reasoning study relies on fixed uniform sampling during inference, overlooking the highly heterogeneous temporal dynamics.
Certain visual attributes, such as illumination conditions and imaging modality, remain relatively stable and are distributed uniformly over time. 
In contrast, critical procedural cues, \eg, transient occlusions, sudden scope movements, or the appearance of specific instruments, often occur sparsely and may only persist for brief moments. As illustrated in Fig.~\ref{fig:vis_policy}, different question types exhibit distinct preferences over temporal sampling patterns,
indicating that a single static sampling policy insufficiently accommodate diverse temporal characteristics.
Motivated by this observation, we propose Temporally-Grounded Multi-Policy Scaling (TMS) to dynamically select sampling policies conditioned on queries. 

TMS first identifies temporally relevant windows and leverages a lightweight policy instructor to determine the most suitable sampling distribution for each window.
Given a video $V$$=$$\{f_t\}_{t=1}^{T}$ and a query $q$, TMS construct a compact yet evidence-dense resampled sequence $V^{*}$ that preserves task-critical temporal cues. We augment the query with an additional instruction:
\emph{``Please estimate the possible happening
time intervals for questions in the video.''}
The model outputs a set of predicted intervals, from which we extract their centers $\{\mu_m\}$ as temporal anchors to construct query-relevant temporal windows 
$\mathcal{B}=\cup_{m} B(\mu_m, w)$:
\begin{equation}
{\small
B(\mu_m, w)=\{t\in\{1,\dots,T\}:\, |t-\mu_m|\leq w\}.
}
\label{eq:bmu}
\end{equation}
Within each window $B_m$, we maintain a candidate policy set 
$\mathcal{P}=\{p_1,\dots,p_K\}$ representing different temporal sampling patterns (\ie, Gaussian, Uniform, Dense, U-shape). 
Rather than manually selecting a fixed policy, a policy instructor generates appropriate proposals conditioned on the window context and the query:
\begin{equation}
{\small
\hat{p} = \operatorname{Instructor}(B_m, q).
}
\label{eq:instructor}
\end{equation}
Given $B_m$, the selected policy $\hat{p}$ shapes the resampling distribution to produce a compact sequence $V^{*}$, which is then fed into the vision--language model for answer generation.
The resampled sequence is then fed into the vision--language model for the final answer generation.
The overall training objective consists of:
\begin{equation}
{\small
\mathcal{L}
=
\mathcal{L}_{\text{QA}}
+
\lambda_{\text{ret}} \mathcal{L}_{\text{ret}}
+
\lambda_{\text{policy}} \mathcal{L}_{\text{policy}}.
}
\end{equation}
Here, $\mathcal{L}_{\text{QA}}$ denotes the next-token prediction loss computed on the final answer generated from $V^{*}$.
The instructor is trained via cross-entropy over the $K$ candidate strategies to optimize $\mathcal{L}_{\text{policy}}$.
The coefficients $\lambda_{\text{ret}}$ and $\lambda_{\text{policy}}$ balance the contributions of retrieval alignment and policy supervision, respectively.

\section{Experiments}
\subsection{Datasets and Evaluation Metrics}
We evaluate \mymodel~on the restructured long-horizon benchmark Colon-LQA and REAL-Colon-VQA~\cite{surgvivqa}. Specifically, we stitch consecutive colonoscopy clips to form long, continuous videos and aggregate their associated question–answer pairs into extended VideoQA samples.
Each long-video instance spans approximately 34 seconds, posing challenges for long-range retrieval and reasoning.
Following previous studies~\cite{surgvivqa}, we evaluate our model using BLEU-4, ROUGE-L, METEOR, and Keyword Accuracy (K-ACC), and report both in-template and out-of-template results. All ablation studies are conducted on Colon-LQA to evaluate each component’s contribution to long-horizon VideoQA.

\subsection{Implementation Details}
All experiments are conducted on two NVIDIA RTX 4090 GPUs. 
The model is trained for three epochs using the AdamW~\cite{adamw} optimizer with a weight decay of 0.01 and a base learning rate of $5$$\times$$10^{-4}$. 
The batch size is 2 with gradient accumulation over 8 steps. 
The visual encoder is initialized from SigLIP2-Large 300M~\cite{sigclip2}, and the language backbone adopts the pre-trained Qwen3-2B~\cite{qwen} model. For policy selection, we use a lightweight instructor built upon BERT-base-uncased~\cite{bert}. 
We employ LoRA~\cite{lora} with rank 8 and scaling factor 16 for parameter-efficient adaptation of the language backbone while keeping its original weights frozen. 
Other task-specific modules are optimized during training.

\begin{table}[t!]
\centering
\scriptsize
\renewcommand{\arraystretch}{1.0}
\setlength{\tabcolsep}{5.0pt}
\caption{Performance comparison on the Colon-LQA benchmark. $^{\dagger}$ denotes zero-shot evaluation without task-specific fine-tuning.}

\begin{tabular}{@{}lcccccccc@{}}
\toprule

\multirow{2.5}{*}{\textbf{Model}}
& \multicolumn{4}{c}{\textbf{In-template}}
& \multicolumn{4}{c}{\textbf{Out-of-template}} \\
\cmidrule(lr){2-5}\cmidrule(lr){6-9}
&
BLE-4 & ROU-L & MET & K-ACC
& BLE-4 & ROU-L & MET & K-ACC \\
\midrule

Qwen3VL$^{\dagger}$~\cite{qwen3vl}
& 6.26 & 31.18 & 29.73 & 41.42
& 2.11 & 29.84 & 26.27 & 38.35 \\

InternVL3$^{\dagger}$~\cite{internvl3}
& 4.37 & 28.05 & 34.62 & 33.18
& 2.64 & 27.57 & 32.88 & 40.12 \\

SurgicalGPT~\cite{Seenivasan2023SurgicalGPT}
& 13.26 & 44.13 & 48.84 & 47.58
& 9.74 & 39.27 & 42.05 & 38.46 \\

PitVQA~\cite{pitvqa}
& 24.73 & 51.78 & 55.06 & 44.17
& 18.67 & 45.18 & 47.73 & 36.28 \\

Qwen3VL~\cite{qwen3vl}
& \underline{63.87} & 72.94 & 74.24 & \underline{63.46}
& \underline{30.47} & \underline{60.14} & \underline{61.57} & \underline{53.18} \\

MedGemma~\cite{medgemma}
& 24.68 & 48.83 & 50.74 & 41.86
& 16.97 & 40.26 & 42.16 & 34.37 \\

InternVL3~\cite{internvl3}
& 57.28 & \underline{73.36} & 72.18 & 60.25
& 27.53 & 56.47 & 57.73 & 49.63 \\

VideoLLaMA3~\cite{videollama3}
& 60.17 & 70.26 & \underline{75.73} & 61.64
& 29.27 & 58.64 & 59.97 & 51.06 \\

SurgViVQA~\cite{surgvivqa}
& 52.46 & 66.74 & 68.17 & 54.63
& 25.86 & 52.47 & 53.66 & 46.94 \\

\textbf{\mymodel}
& \textbf{74.37} & \textbf{85.03} & \textbf{83.00} & \textbf{66.67}
& \textbf{38.01} & \textbf{69.50} & \textbf{65.85} & \textbf{65.73} \\

\bottomrule
\end{tabular}

\label{tab:colon_lqa_results}
\end{table}

\subsection{Comparison with State-of-the-art Methods}
Table~\ref{tab:colon_lqa_results} and Table~\ref{tab:real_colon_vqa_results} compare our method with state-of-the-art approaches on the Colon-LQA long-video benchmark and REAL-Colon-VQA~\cite{surgvivqa}. Due to the poor zero-shot performance observed, we fine-tune all compared methods under a unified task-specific training protocol (same splits, resolution, and metrics) following their default implementations to ensure fair and meaningful evaluation. 

\begin{figure}[t]
    \centering
    \begin{tabular}{c}
        \includegraphics[width=0.99\textwidth]{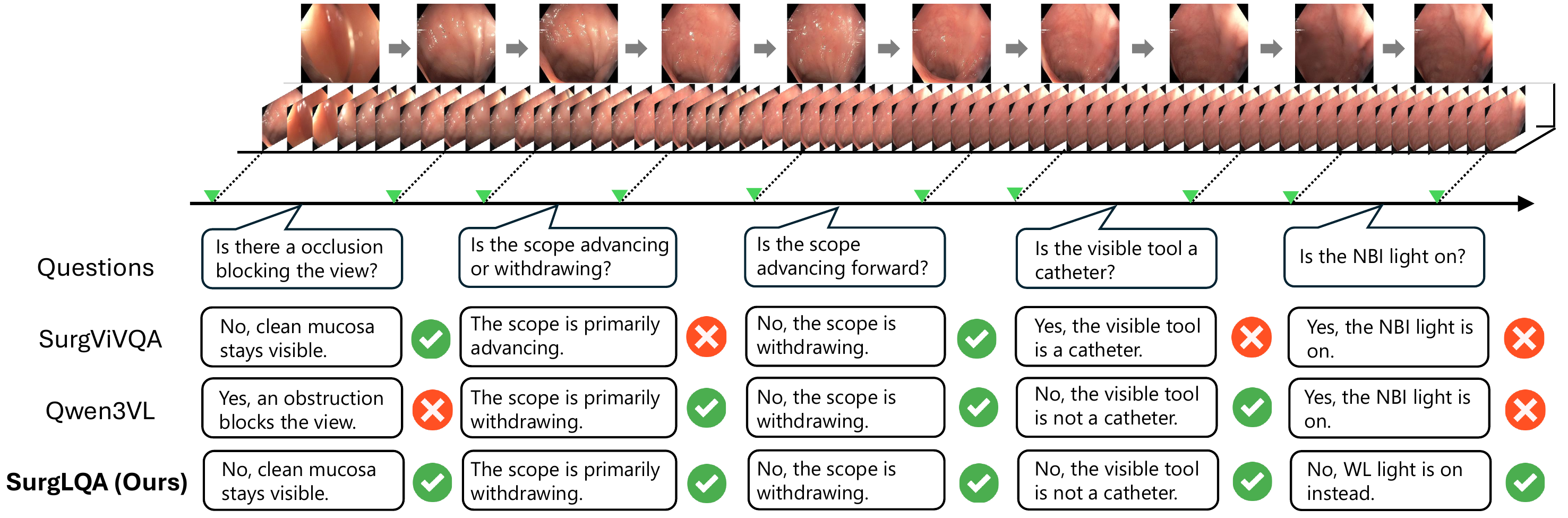} 
    \end{tabular}
    \caption{Qualitative comparison with representative models on Colon-LQA.}
\label{fig:qualitative_results} 

\end{figure}

On Colon-LQA long-video bench, our model achieves the superior performance across all four metrics, highlighting its long-range understanding capabilities. 
Under out-of-template setting, where question formulations differ from those seen during training, our method maintains clear advantages, suggesting deeper semantic reasoning rather than reliance on template-specific cues. 
Fig.~\ref{fig:qualitative_results} further demonstrates improved temporal coherence and contextual consistency over extended video sequences. 
On REAL-Colon-VQA, while the benefit of long-horizon modeling is naturally reduced, 
our method still achieves competitive performance, indicating strong generalization to standard short-video benchmarks.

\begin{table}[t!]
\centering
\scriptsize
\renewcommand{\arraystretch}{1.0}
\setlength{\tabcolsep}{5.0pt}
\caption{
Performance comparison on the REAL-Colon-VQA~\cite{surgvivqa} benchmark.
$^{\dagger}$ denotes zero-shot evaluation without task-specific fine-tuning.
}

\begin{tabular}{@{}lcccccccc@{}}
\toprule

\multirow{2.5}{*}{\textbf{Model}}
& \multicolumn{4}{c}{\textbf{In-template}}
& \multicolumn{4}{c}{\textbf{Out-of-template}} \\
\cmidrule(lr){2-5}\cmidrule(lr){6-9}
&
BLE-4 & ROU-L & MET & K-ACC
& BLE-4 & ROU-L & MET & K-ACC \\
\midrule

Qwen3VL$^{\dagger}$~\cite{qwen3vl}
& 7.31 & 40.34 & 44.87 & 45.40
& 6.42 & 36.42 & 41.19 & 47.86 \\

InternVL3$^{\dagger}$~\cite{internvl3}
& 7.13 & 43.90 & 54.31 & 68.67
& 3.87 & 32.10 & 44.34 & 53.73 \\

SurgicalGPT~\cite{Seenivasan2023SurgicalGPT}
& 14.93 & 47.85 & 52.36 & 33.33
& 12.35 & 42.87 & 50.49 & 46.67 \\

PitVQA~\cite{pitvqa}
& 64.55 & 78.48 & 79.99 & 54.13
& 23.63 & 50.03 & 53.22 & 42.93 \\

Qwen3VL~\cite{qwen3vl}
& \underline{82.40} & \underline{88.75} & \underline{89.10} & \underline{72.35}
& 35.82 & \underline{64.50} & \underline{63.41} & 66.28 \\

MedGemma~\cite{medgemma}
& 69.85 & 80.92 & 82.40 & 60.78
& 27.16 & 52.80 & 50.43 & 48.62 \\

InternVL3~\cite{internvl3}
& 72.30 & 83.15 & 85.22 & 65.91
& 29.44 & 55.36 & 53.90 & 52.84 \\

VideoLLaMA3~\cite{videollama3}
& 76.47 & 85.60 & 84.75 & 68.34
& \underline{37.08} & 59.82 & 58.55 & \underline{66.97} \\

SurgViVQA~\cite{surgvivqa}
& 71.98 & 82.85 & 84.11 & 63.20
& 31.19 & 53.62 & 54.89 & 47.73 \\

\textbf{\mymodel}
& \textbf{87.69} & \textbf{92.93} & \textbf{91.85} & \textbf{79.60}
& \textbf{44.89} & \textbf{72.98} & \textbf{69.59} & \textbf{78.93} \\

\bottomrule
\end{tabular}

\label{tab:real_colon_vqa_results}
\end{table}

\subsection{Ablation Study}
\begin{figure}[t]
  \centering
  \begin{minipage}{0.56\linewidth}
    \centering
    \includegraphics[width=\linewidth]{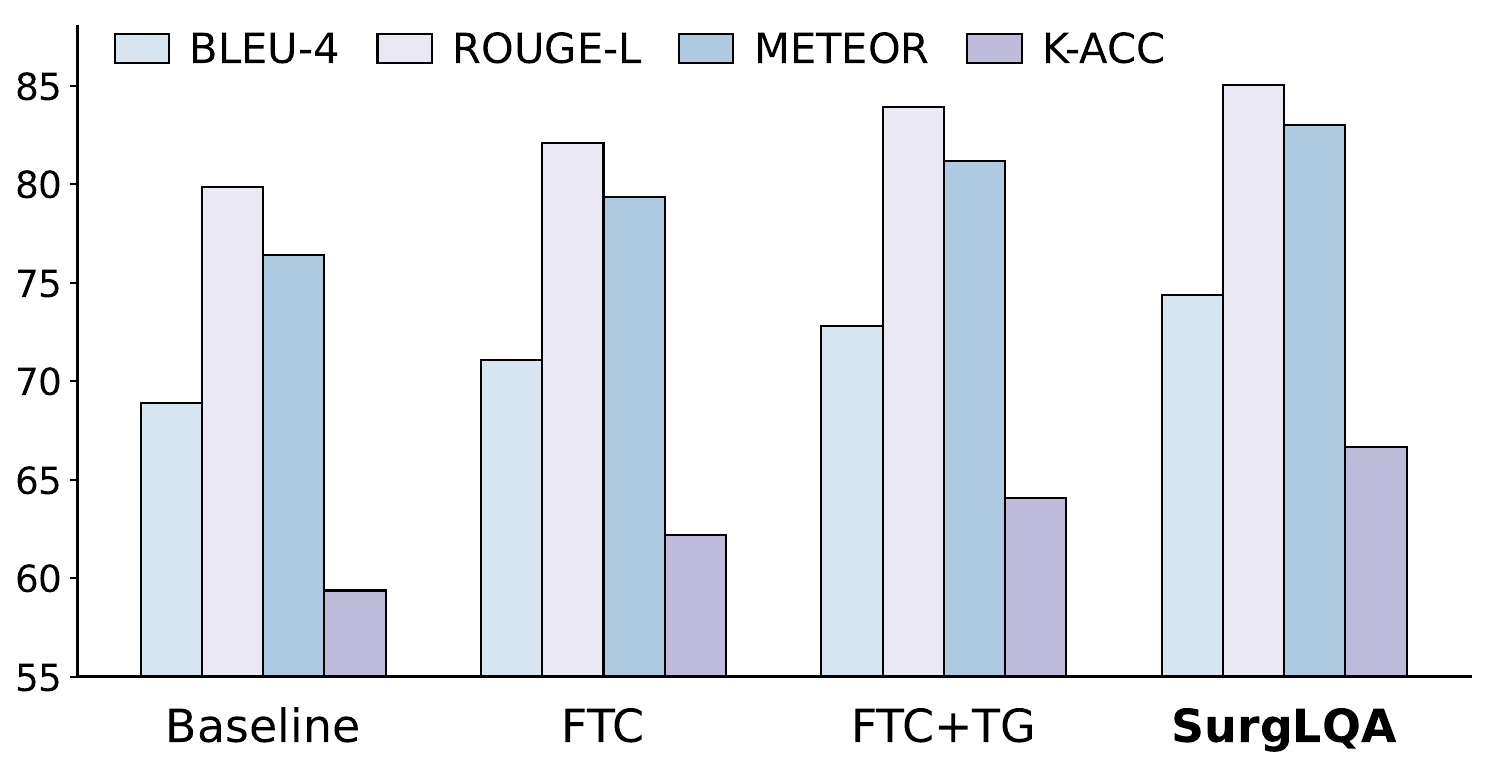}
  \end{minipage}\hfill
  \begin{minipage}{0.38\linewidth}
    \centering
    \includegraphics[width=\linewidth]{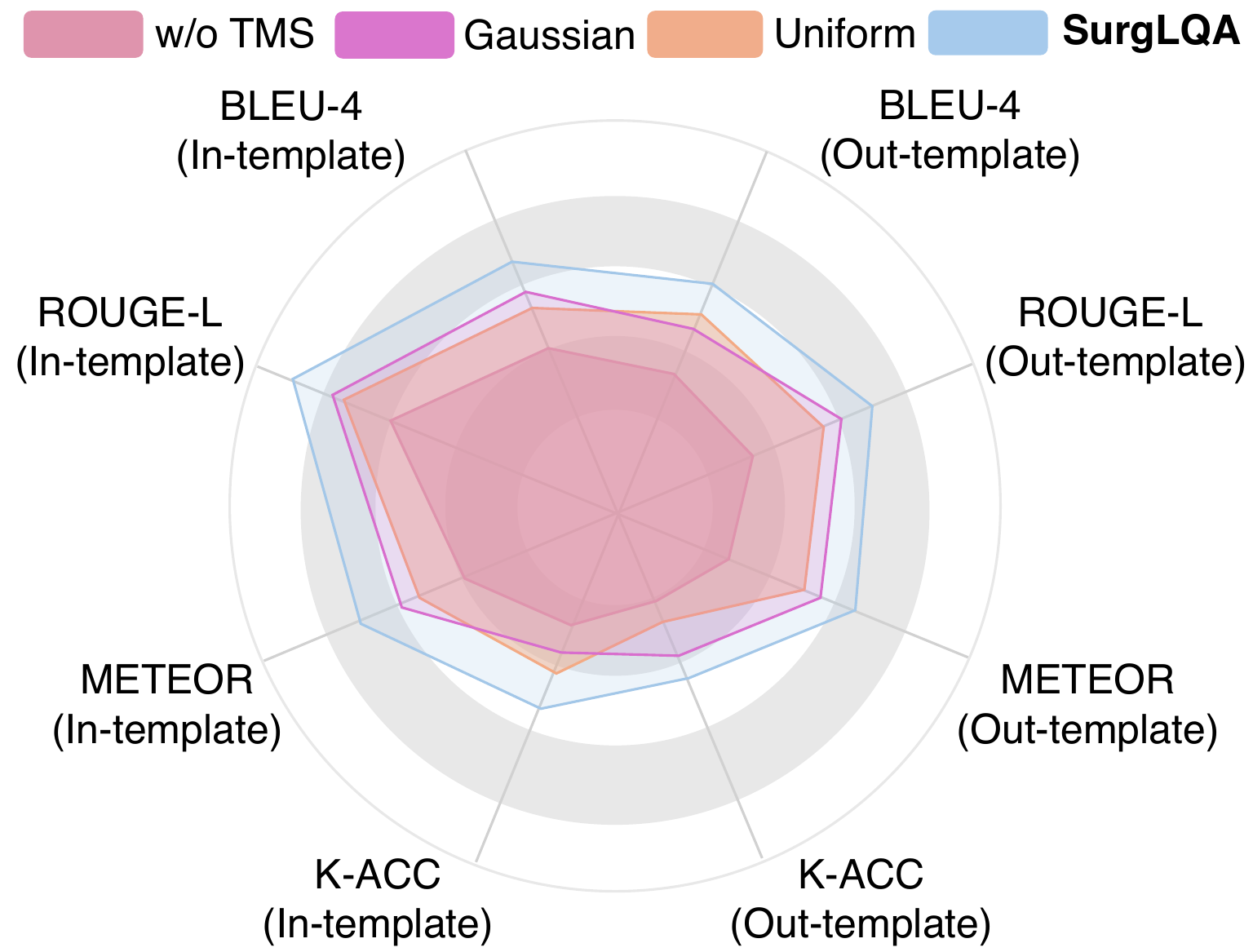}
  \end{minipage}

  \caption{Left: Ablations of key components in \mymodel. Right: Ablations of TMS.}
  \label{fig:ablation_combined}
\end{figure}

\noindent\textbf{Contributions of Key Components.}
As shown in the left of \figref{fig:ablation_combined}, introducing FTC consistently improves all metrics, indicating more effective preservation of critical temporal cues with reduced redundancy. 
Adding Temporal Grounding (TG) further enhances performance, particularly in keyword accuracy, reflecting the benefit of aligning temporal sampling with query-relevant intervals. 
With Policy Instructor (PI), the model achieves better overall results, demonstrating the advantage of adaptive test-time policy selection over fixed strategies. 

\begin{figure}[t!]
    \centering
    \begin{tabular}{c}
        \includegraphics[width=0.98\textwidth]{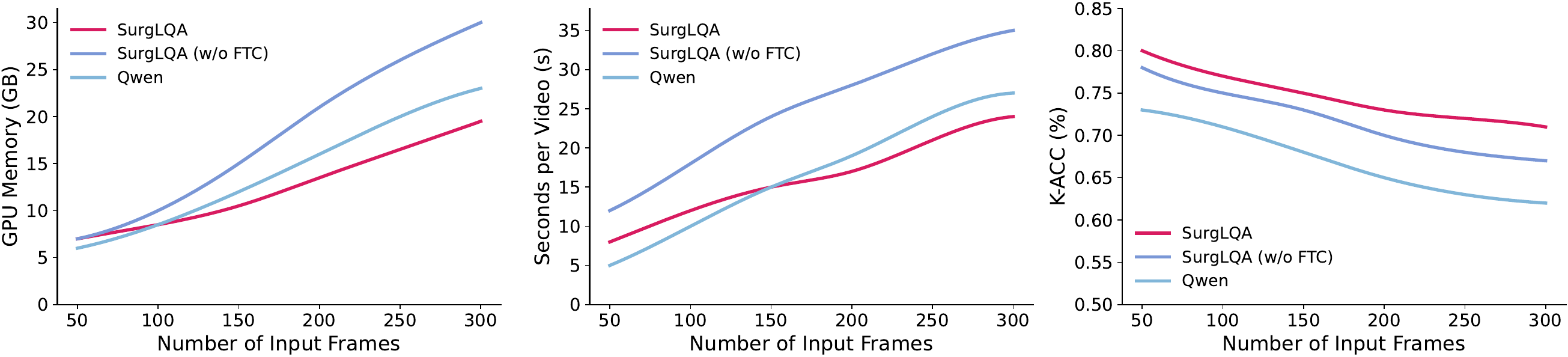} 
    \end{tabular}
    \caption{Scalability analysis with increasing input frames.}
\label{fig:fps_analysis} 
\end{figure}
\noindent\textbf{Ablations for TMS.}
The right part of \figref{fig:ablation_combined} compares different inference strategies within TMS.
As shown in the figure, removing TMS yields the weakest performance, while incorporating temporal grounding with a single policy improves all metrics, demonstrating the benefit of focusing on temporally relevant intervals.
The better results are achieved with full policy selection,
indicating that adaptive multi-policy scaling further enhances long-range reasoning.

\noindent\textbf{Scalability Analysis.}
As illustrated in Fig.~\ref{fig:fps_analysis}, removing FTC substantially increases memory and runtime while reducing K-ACC as the frame budget grows. 
Direct frame-level scaling thus incurs significant computational overhead without effectively preserving temporal evidence. 
In contrast, \mymodel~maintains lower resource consumption and consistently higher K-ACC, confirming the effectiveness of temporally faithful compression for efficient long-horizon reasoning.

\section{Discussion}
\label{sec:discussion}

\noindent\textbf{Clinical Relevance and Deployment.} 
The practical utility of surgical VideoQA frameworks heavily depends on their clinical relevance and computational feasibility in real-time intraoperative environments. Combined with the runtime efficiency analysis illustrated in \figref{fig:fps_analysis}, SurgLQA achieves low-latency inference and exceptional computational scalability under extended temporal horizons, strongly supporting its potential for seamless integration into practical clinical workflows and real-time intraoperative decision assistance.

\noindent\textbf{Beyond Counterparts.} 
Unlike generic long-video methods that miss sparse, fine-grained surgical events via uniform sampling or token compression, SurgLQA targets long-horizon sparse evidence retrieval. FTC preserves temporal fidelity via a dedicated retrieval loss, while TMS replaces rigid grids with query-conditioned grounding and multi-policy re-sampling to capture critical details.


\section{Conclusion}
 
This paper introduces SurgLQA, a unified long-horizon framework for surgical video question answering. 
Faithful Temporal Consolidation models extended surgical procedures by constructing compact yet temporally coherent representations that preserve fine-grained procedural cues. 
Temporally-Grounded Multi-Policy Scaling further regulates reasoning granularity across temporally salient contexts, enabling grounded modeling of temporal dynamics within long surgical workflows. 
Experimental results on the restructured long-duration benchmark Colon-LQA and REAL-Colon-VQA demonstrate consistent improvements in long-range reasoning enabled by event-level temporal grounding.
\bigskip

\noindent\textbf{Acknowledgment.}
This work was supported by the Research Grants Council of the Hong Kong Special Administrative Region, China, under Project T45-401/22-N, and by the Hong Kong Innovation and Technology Fund, under Project GHP/252/23SZ.

%
%
    \small
    \bibliographystyle{splncs04}
    \bibliography{ref}

\end{document}